\pdfoutput=1
\documentclass[11pt]{article}
\usepackage{authblk}
\usepackage{setspace}
\usepackage{booktabs}
\usepackage{ACL2023}
\usepackage{setspace}
\usepackage{times}
\usepackage{latexsym}
\usepackage{graphicx}
\usepackage{multirow, adjustbox}
\usepackage{xcolor,colortbl}
\usepackage{booktabs}
\usepackage{arydshln}
\usepackage{graphicx}
\usepackage[T1]{fontenc}
\usepackage[utf8]{inputenc}

\usepackage{microtype}

\usepackage{inconsolata}
\usepackage{arabtex}
\renewenvironment{abstract}%
         {\centerline{\large\bf Abstract}%
          \begin{list}{}%
             {\setlength{\rightmargin}{0.6cm}%
              \setlength{\leftmargin}{0.6cm}}%
           \item[]\ignorespaces}%
         {\unskip\end{list}}
\usepackage{utf8}
\setcode{utf8}
\title{FaBERT: Pre-training BERT on Persian Blogs}

\author[$\diamondsuit$$\dagger$]{Mostafa Masumi}
\author[$\diamondsuit$]{Seyed Soroush Majd}
\author[$\diamondsuit$]{Mehrnoush Shamsfard}
\author[$\dagger$]{Hamid Beigy}
\affil[$\diamondsuit$]{Computer Science and Engineering Department, Shahid Beheshti University}
\affil[$\diamondsuit$]{\textit {s.majd@mail.sbu.ac.ir, m-shams@sbu.ac.ir}}
\affil[$\dagger$]{Computer Engineering Department, Sharif University of Technology}
\affil[$\dagger$]{\textit {\{m.masumi, beigy\}@sharif.edu}}

\begin{document}

\maketitle

\begin{abstract}
We introduce FaBERT, a Persian BERT-base model pre-trained on the HmBlogs corpus, encompassing both informal and formal Persian texts. FaBERT is designed to excel in traditional Natural Language Understanding (NLU) tasks, addressing the intricacies of diverse sentence structures and linguistic styles prevalent in the Persian language. In our comprehensive evaluation of FaBERT on 12 datasets in various downstream tasks, encompassing Sentiment Analysis (SA), Named Entity Recognition (NER), Natural Language Inference (NLI), Question Answering (QA), and Question Paraphrasing (QP), it consistently demonstrated improved performance, all achieved within a compact model size. The findings highlight the importance of utilizing diverse and cleaned corpora, such as HmBlogs, to enhance the performance of language models like BERT in Persian Natural Language Processing (NLP) applications. FaBERT is openly accessible at \url{https://huggingface.co/sbunlp/fabert}.
\end{abstract}

\section{Introduction}

In recent times, we've seen the rise of sophisticated language models like BERT \cite{devlin2018bert}, transforming the understanding of languages, including Persian. Whether designed for multiple languages or specifically for Persian, these models have been employed across various applications in Persian Natural Language Processing (NLP). Their training encompassed a diverse range of textual sources, including websites like Wikipedia and social media platforms such as Twitter, as well as news articles and academic journals.

More recently, Large Language Models (LLMs) with a substantial increase in parameters, have significantly reshaped the landscape of NLP, excelling in a myriad of tasks. Despite their significant contributions, finely-tuned LMs such as BERT still demonstrate robust performance, achieving comparable results or, in many cases, even outperforming LLMs in traditional Natural Language Understanding (NLU) tasks, including Natural Language Inference (NLI), Sentiment Analysis, Text Classification, and Question Answering (QA) \cite{yang2023harnessing}.

Additionally, LLMs often come with the drawback of slower response times and increased latency compared to smaller models. Moreover, the use of LLMs typically demands advanced hardware, creating accessibility challenges for many users. Privacy concerns may also emerge when employing LLMs online. On the other hand, smaller LMs are more suitable for use in local standard computers and settings with limited computing capabilities, given their compact design.  

Our motivation is to develop FaBERT, a Persian BERT-base model, to enhance performance in traditional NLU tasks and enable efficient processing of both formal and informal texts in the language. While existing Persian LMs exhibit commendable capabilities, there remains room for improvement, especially in handling the complexities of Persian informal texts. Informal Persian in real-world communication has its unique features like flexible sentence structures, cultural references, informal lexicon, and slang. FaBERT is designed to tackle these potential challenges and improve overall performance.

Our findings reveal that the cleaned corpus from Persian blogs enhances the model's performance, leading to state-of-the-art results across various downstream tasks. The main contributions of this paper are:
\begin{enumerate}
\item	Pre-training a BERT-base model on Persian blog texts in the HmBlogs corpus, and making it publicly accessible.
\item	Evaluating the model's performance on 12 datasets in various downstream tasks, including sentiment analysis, irony detection, natural language inference, question paraphrasing, named entity recognition, and question answering.
\end{enumerate}
The subsequent sections of the paper are structured as follows: Section 2 provides an introduction and comparison of various BERT models employed for Persian NLP. Section 3 delves into the details of our corpus, model, and its pre-training procedure. Section 4 compares FaBERT's performance in downstream tasks with other models. Finally, Section 5 concludes the paper by summarizing our findings.

%-------------------------------------------------------------------------------------------------------
\section{Related Works}\label{sec:related_work}
BERT, which stands for Bidirectional Encoder Representations from Transformers, has demonstrated its exceptional abilities across a wide range of natural language understanding tasks. Unlike traditional language models that process text in a unidirectional manner (left-to-right or right-to-left), BERT considers both the left and right context of words.

BERT's pre-training involved two training objectives: Masked Language Modeling (MLM) and Next Sentence Prediction (NSP). MLM randomly masks words in a sentence, and the model learns to predict the missing words based on context, enhancing its ability to grasp the semantic meaning and relationships between words within sentences. On the other hand, in the NSP task, the model has to predict whether sentence B logically succeeds sentence A.

MLM and NSP are designed for the model to learn a language representation, which can then be used to extract features for downstream tasks. Continuing the discussion, we will present a selection of Persian-language BERT models.

The most well-known Persian language model is ParsBERT \cite{farahani2021parsbert}. It was pre-trained using both MLM and NSP tasks, utilizing a training corpus collected from 8 different sources. ParsBERT has become the preferred choice for Persian NLP tasks, thanks to its outstanding performance. Ariabert \cite{ghafouri2023ariabert} is another Persian language model that follows RoBERTa's enhancements \cite{liu2019roberta} and utilizes Byte-Pair Encoding tokenizer. It's diverse training dataset, exceeding 32 gigabytes, includes conversational, formal, and hybrid texts.

Additionally, many Multilingual Language Models have been released since, and few of them include Persian. Multilingual BERT, also known as mBERT, was introduced by \cite{devlin2018bert}. It was trained with NSP and MLM tasks on the Wikipedia pages of 104 languages with a shared word-piece vocabulary. mBERT has shown impressive zero-shot cross-lingual transfer and is effective in utilizing task-specific annotations from one language for fine-tuning and evaluation in another. Although mBERT has shown solid performance across different languages, monolingual BERT models outperform mBERT in most downstream tasks.

Similarly, XLM-R \cite{conneau2019unsupervised}, an extension of the RoBERTa model by Facebook AI, is designed for cross-lingual understanding. This model was pre-trained with the MLM objective on a vast corpus comprising more than 2 terabytes of text from 100 languages and outperformed mBERT in many downstream tasks.

The models previously reviewed adhere to the architecture introduced by the original BERT-base model, featuring 12 layers and 12 attention heads. While maintaining this consistency, there are variations in vocabulary size among these models.

A larger vocabulary facilitates the capture of more unique tokens and their relationships, but it comes at the expense of increased parameters. This, in turn, necessitates more extensive training data for learning embeddings. Conversely, smaller vocabularies may struggle to capture all the details of language, potentially causing information and context to be lost.

An instance is found in the multilingual model mBERT, which supports 100 different languages with a vocabulary size of only 100,000. Despite the broad language coverage, this choice leads to a limited set of tokens for each language. Consequently, sentences are transformed into a greater number of tokens, potentially exceeding the maximum supported sequence length and resulting in the loss of information. Table \ref{tab:Vocab_Parameter_Comparison} summarizes the vocabulary size and number of parameters for each model under consideration.

\begin{table}[!htp]
\centering
\begin{adjustbox}{width={0.5\textwidth},totalheight={\textheight},keepaspectratio}
\begin{tabular}{lccc}
\toprule
\textbf{Model} & \textbf{Vocabulary Size (K)} & \textbf{\# of Parameters (M)} \\
\midrule
    BERT (English) & 30 & 109 \\
    mBERT & 105 & 167 \\
    XLM-R & 250 & 278 \\
    ParsBERT & 100 & 162 \\
    %\rowcolor{lightgray}
    AriaBERT & 60 & 132 \\

\bottomrule
\end{tabular}
\end{adjustbox}
\captionsetup{font=footnotesize}
\caption{Vocabulary Size and Parameter Count of Persian BERT Models}\label{tab:Vocab_Parameter_Comparison}
\end{table}

%nice

\section{Methodology}\label{sec:methodology}
\subsection{Training Corpus}
The selection of an appropriate training corpus is a pivotal element in the pre-training of a language model. For this effort, we utilized the HmBlogs corpus \cite{khansari2021hmblogs}, a collection of 20 million posts of Persian blogs over 15 years. HmBlogs includes more than 6.8 billion tokens, covering a wide range of topics, genres, and writing styles, including both formal and informal texts together. 

To ensure high-quality pre-training, a series of pre-processing steps were performed on the corpus. Many posts written in the Persian alphabet were erroneously identified as Persian despite not being in the Persian language. This confusion arises from the Persian alphabet's resemblance to the alphabets of other languages like Arabic and Kurdish. Additionally, some other posts had typographical errors, very rare words, or the excessive use of local dialects. Therefore, a post-discriminator was implemented to filter out these improper and noisy posts. 
Cleaning documents in Persian poses another challenge due to the presence of non-standard characters\footnote{For instance, Arabic '\raisebox{0.3ex}{\<\small{ي}>}' and '\raisebox{0.3ex}{\<\small{ك}>}' are occasionally substituted for Persian '\raisebox{0.6ex}{\<\small{ی}>}' and '\raisebox{0.6ex}{\<\small{ک}>}'.}.
These characters look identical to Persian characters, but their different codes can cause problems during pre-training. Some Persian blogs may also use decorative characters to make the text visually appealing. Such characters were standardized to ensure uniform representation and avoid potential discrepancies. Additionally, numbers were replaced with a default value, and words with repetitive characters were corrected.

\subsection{Pre-training Procedure}
We trained a BERT-base model similar to that proposed by \cite{devlin2018bert}. Our BERT-base model, FaBERT, replicates the original architecture with 12 hidden layers, each comprising 12 self-attention heads.

We opted for the WordPiece tokenizer over alternatives such as BPE, as prior evidence indicates no performance improvement \cite{geiping2023cramming}, and with a conservative stance, we set the vocabulary size to 50,000 tokens. This decision aimed at finding a balance between capturing linguistic details and managing the computational demands associated with larger vocabularies. It's essential to note that Persian text includes half spaces, a feature absent in English. Consequently, the FaBERT tokenizer has been adapted to handle this feature, ensuring appropriate representation of texts during pre-training and fine-tuning.

The total number of parameters for FaBERT is 124 million. In comparison to other Persian and multilingual base models outlined in Table \ref{tab:Vocab_Parameter_Comparison}, FaBERT is more compact with fewer parameters.

During pre-training, each input consisted of one or more sentences sampled contiguously from a single document. The samples were of varying lengths to help the model effectively learn the positional encodings. 

We implemented dynamic masking, inspired by the methodology introduced in \cite{liu2019roberta}, and omitted the Next Sentence Prediction task from our pre-training process, as it was demonstrated to have no discernible positive impact on performance. The masking rate for dynamic masking was set to 15\%. We also utilized the whole word masking approach for enhanced performance. Unlike traditional MLM, which randomly masks individual tokens in a sentence, whole word masking involves masking entire words. Table \ref{tab:Fabert_Hyperparameters} details the hyperparameters used in the pre-training process.

\begin{figure}
\centering
\includegraphics[width=\linewidth]{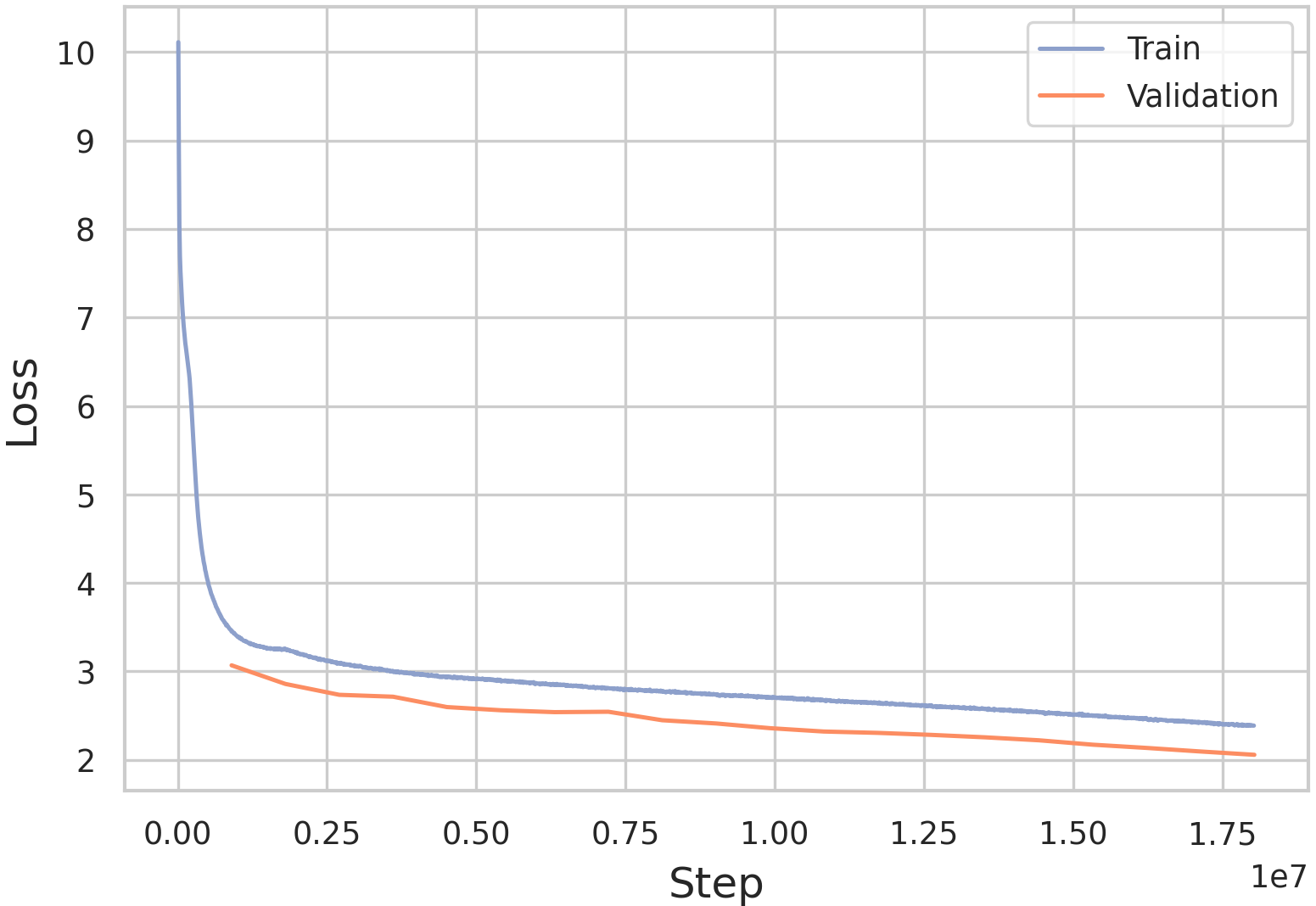}
\caption{Train and Validation MLM loss in pre-training}
\label{fig:train}
\end{figure}

The training was conducted on a single Nvidia A100 40GB GPU, spanning a duration of 400 hours. The final validation perplexity achieved was 7.76, and the train and validation loss plot is presented in Figure \ref{fig:train}.

\begin{table}
\centering
\begin{adjustbox}{width={0.5\textwidth},totalheight={\textheight},keepaspectratio}
\begin{tabular}{lc}
\hline
\textbf{Hyperparameter} & \textbf{Value}\\
\hline
Batch Size & 32 \\
Optimizer & Adam \\
Learning Rate & 6e-5 \\
Weight Decay & 0.01 \\
\hline
\end{tabular}
\begin{tabular}{lc}
\hline
\textbf{Hyperparameter} & \textbf{Value}\\
\hline
Total Steps & 18 Million \\
Warmup Steps & 1.8 Million \\
Precision Format & TF32 \\
Dropout & 0.1 \\
\hline
\end{tabular}
\end{adjustbox}
\caption{Pre-training Hyperparameters}
\label{tab:Fabert_Hyperparameters}
\end{table}

%-----------------------------------------------------------------------------------
\section{Experiments and Results}\label{sec:experiments_and_results}
In this section, we assess the FaBERT model across four different categories of downstream tasks. For NLI and Question Paraphrasing, sentence pairs are processed to generate labels based on their relationship. In NER, entities within single input sentences are labeled at the token level. Sentiment Analysis and Irony Detection involve processing individual sentences and assigning corresponding labels. In Question Answering, models utilize a given question and the provided paragraph to generate token-level spans for answers. Lastly, we analyze the efficiency of FaBERT’s tokenizer and compare it with other BERT models.

In fine-tuning each dataset, a grid search is employed, utilizing train/validation/test splits. The reported scores correspond to the test set and are based on hyperparameters that yield the best validation scores. The scope of the grid search and the split sizes for each dataset can be found in Appendix \ref{sec:appendix_hyper}.

\subsection{Natural Language Inference and Question Paraphrasing} 
In this section, we analyze FaBERT's ability to understand logical and semantic relationships between sentences, focusing on tasks like Natural NLI and Question Paraphrasing.  We assess its performance using the Farstail \cite{amirkhani2023farstail}, SBU-NLI \cite{rahimi2024knowledge}, and ParsiNLU Question Paraphrasing \cite{khashabi2021parsinlu} datasets.

\subsubsection*{FarsTail}
The FarsTail NLI dataset is sourced from multiple-choice questions from various subjects, specifically collected from Iranian university exams. Each of these questions became the basis for generating NLI instances with three different relationships: Entailment, Contradiction, and Neutral.

\subsubsection*{SBU-NLI}
SBU-NLI is another dataset containing sentence pairs categorized into three labels: Entailment, Contradiction, and Neutral. This data is gathered from various sources to create a balanced dataset.

\subsubsection*{ParsiNLU Question Paraphrasing}
This task involves determining the relationship between pairs of questions, specifically classifying whether they are paraphrases. The dataset is created through two means: first, by mining questions from Google auto-complete and Persian discussion forums, and second, by translating the QQP dataset with Google Translate API. As a result, some questions are presented in an informal fashion.

\begin{table}[htbp]
\centering
\begin{adjustbox}{width={0.5\textwidth},totalheight={\textheight},keepaspectratio}
  \begin{tabular}{lccc}
    \toprule
    \textbf{Model} & \textbf{FarsTail} & \textbf{SBU-NLI} & \textbf{Parsi-NLU QP} \\
    \midrule
    ParsBERT & 82.52 & 58.41 & 77.60 \\
    mBERT & 83.42 & 66.38 & 79.48 \\
    XLM-R & 83.50 & 58.85 & 79.74 \\
    AriaBERT & 76.39 & 52.81 & 78.86 \\
    \midrule
    FaBERT & \textbf{84.45} & \textbf{66.65} & \textbf{82.62} \\
    \bottomrule
  \end{tabular}
\end{adjustbox}
\caption{Performance Comparison in NLI and Question Paraphrasing}\label{tab:NLI_Results}
\end{table}

As observed in Table \ref{tab:NLI_Results}, FaBERT demonstrates a +1\% improvement in F1 for FarsTail, comparable performance to mBERT in SBU-NLI, and a +2.88\% F1 score in the informal ParsiNLU Question Paraphrasing dataset.

\subsection{Named Entity Recognition}
In this section, we assess the efficacy of FaBERT in NER, a commonly employed intermediate task that facilitates information extraction and entity identification within textual data. Our assessment leveraged formal and informal datasets, including ParsTwiNER \cite{aghajani2021parstwiner}, PEYMA \cite{shahshahani2018peyma}, and MultiCoNER v2 \cite{fetahu2023multiconer}. The comparison of different models for each entity type is detailed in Appendix \ref{sec:appendix_ner}.

\subsubsection*{ParsTwiNER}
The ParsTwiNER offers a NER dataset gathered from 7632 tweets collected from the Persian Twitter accounts, offering diverse informal Persian content. Annotation by experts in natural language processing resulted in 24061 named entities across categories such as persons, organizations, locations, events, groups, and nations.

\subsubsection*{PEYMA}
The PEYMA NER dataset, derived from formal text extracted from ten news websites, classifies words into different categories, encompassing persons, locations, organizations, time, date, and more. PEYMA is known as a key asset for training and evaluating NER systems in the Persian language.

\subsubsection*{MultiCoNER v2}
Initially introduced as a part of SemEval task in 2022, MultiCoNER is a multilingual NER dataset crafted to address contemporary challenges in NER, such as low-context scenarios, syntactically complex entities like movie titles, and long-tail entity distributions. The enhanced version of this dataset was used in the following year as part of the SemEval 2023 task. This version, known as MultiCoNER v2, expanded these challenges by adding fine-grained entities and inserting noise in the input text. Gathered from Wikidata and Wikipedia, the dataset spans 12 languages, with Persian being the focus of our evaluations.

\begin{table}[htbp]
\centering
\begin{adjustbox}{width={0.5\textwidth},totalheight={\textheight},keepaspectratio}
  \begin{tabular}{lccc}
    \toprule
    \textbf{Model} & \textbf{ParsTwiner} & \textbf{PEYMA} & \textbf{MultiCoNER v2} \\
    \midrule
    ParsBERT & 81.13 & 91.24 & \textbf{58.09} \\
    mBERT & 75.60 & 87.84 & 51.04 \\
    XLM-R & 79.50 & 90.91 & 51.47 \\
    AriaBERT & 78.53 & 89.76 & 54.00 \\
    \midrule
    FaBERT & \textbf{82.22} & \textbf{91.39} & 57.92 \\
    \bottomrule
  \end{tabular}
\end{adjustbox}
\caption{Performance Comparison in Named Entity Recognition}\label{tab:NER_Results}
\end{table}

The evaluation metrics used include micro-F1 for PEYMA and ParsTwiNER datasets, and macro-F1 for MultiCoNER v2. Table \ref{tab:NER_Results} provides a detailed overview of scores achieved by each model.
Across the board, all models demonstrated comparable performance in the PEYMA dataset. However, FaBERT model exhibited a slight improvement by achieving a +1.09\% increase in F1 score for the informal ParsTwiNER dataset. In the MultiCoNER v2 dataset, both FaBERT and ParsBERT outperformed other models. In general FaBERT and ParsBERT seem to be great options for applications involving NER.

\subsection{Sentiment Analysis and Irony Detection}
In this section, we assess FaBERT's performance in classifying expressions. We employed DeepSentiPers \cite{sharami2020deepsentipers}, MirasOpinion \cite{asli2020optimizing}, and MirasIrony \cite{golazizian2020irony} datasets for evaluation.

\subsubsection*{DeepSentiPers}
The DeepSentiPers dataset comprises 9,000 customer reviews of Digikala, an Iranian E-commerce platform. Originally, each sentence's polarity was annotated using a 5-class label set $E = \{-2, -1, 0, +1, +2\}$, representing sentiments from very displeased to delighted. However, our investigation revealed inconsistencies, particularly between the -1 and -2 categories for negative sentiments and the +1 and +2 categories for positive sentiments. Recognizing the overlap between these closely related labels, we opted for a simplified 3-class labeling approach, classifying sentiments as negative, neutral, or positive.

\subsubsection*{MirasOpinion}
MirasOpinion, the largest Persian Sentiment dataset, comprises 93,000 reviews gathered from the Digikala platform. Through crowdsourcing, each review was labeled as Positive, Neutral, or Negative. This dataset was included in the SPARROW, a benchmark for sociopragmatic meaning understanding. Participating in the SPARROW benchmark \cite{zhang2023skipped} allowed us to assess FaBERT against various language models. 

\subsubsection*{MirasIrony}
MirasIrony, a 2-labeled dataset designed for irony detection, encompasses 4,339 manually labeled Persian tweets. In this dataset, tweets exhibiting a disparity between their literal meaning and sentiment were labeled as positive, while those lacking this characteristic were labeled as negative. Similar to MirasOpinion, we assessed the performance of models on MirasIrony using the SPARROW benchmark.

\begin{table}[htbp]
\centering
\begin{adjustbox}{width={0.5\textwidth},totalheight={\textheight},keepaspectratio}
  \begin{tabular}{lccc}
    \toprule
    \textbf{Model} & \textbf{DeepSentiPers} & \textbf{MirasOpinion} & \textbf{MirasIrony} \\
    \midrule
    ParsBERT & 74.94 & 86.73 & 71.08 \\
    mBERT & 72.95 & 84.40 & 74.48 \\
    XLM-R & 79.00 & 84.92 & \textbf{75.51} \\
    AriaBERT & 75.09 & 85.56 & 73.80 \\
    \midrule
    FaBERT & \textbf{79.85} & \textbf{87.51} & 74.82 \\
    \bottomrule
  \end{tabular}
\end{adjustbox}
\caption{Performance Comparison in Sentiment Analysis and Irony Detection}\label{tab:Sentiment_Irony_Results}
\end{table}

Macro averaged F1 score serves as the evaluation metric for DeepSentiPers and MirasOpinion, while Accuracy is employed for MirasIrony. As presented in Table \ref{tab:Sentiment_Irony_Results}, FaBERT achieved the highest scores in sentiment analysis for both DeepSentiPers and MirasOpinion. For irony detection in the MirasIrony dataset, XLM-R outperforms other models, securing the leading position with a score of 75.51\%. FaBERT demonstrated notable performance as well, securing the second spot with 74.8\% accuracy. Through the SPARROW benchmark leaderboard, other models can be compared with FaBERT on MirasOpinion\footnote{\url{https://sparrow.dlnlp.ai/sentiment-2020-ashrafi-fas.taskshow}} and MirasIrony\footnote{\url{https://sparrow.dlnlp.ai/irony-2020-golazizian-fas.taskshow}} tasks.

\begin{table*}[!htbp]
\centering
\begin{adjustbox}{width={\textwidth},totalheight={\textheight},keepaspectratio}
\begin{tabular}{lcccccccccccc}
\toprule
\multirow{2}{*}{\textbf{Model}} & \multicolumn{2}{c}{\textbf{ParsiNLU}} & \multicolumn{5}{c}{\textbf{PQuAD}} & \multicolumn{5}{c}{\textbf{PCoQA}} \\
\cmidrule(lr){2-3} \cmidrule(lr){4-8} \cmidrule(lr){9-13}
& \textbf{Exact Match} & \textbf{F1} & \textbf{Exact Match} & \textbf{F1} & \textbf{HasAns EM} & \textbf{HasAns F1} & \textbf{NoAns} & \textbf{Exact Match} & \textbf{F1} & \textbf{HEQ-Q} & \textbf{HEQ-M}  & \textbf{NoAns}\\
\midrule
ParsBERT & 22.10 & 44.89 & 74.41 & 86.89 & 68.97 & 85.34 & 91.79 & 31.17 & 50.96 & 41.07  & 0.81 & 48.83\\
mBERT & 26.31 & 49.63 & 73.68 & 86.71 & 67.52 & 84.66 & \textbf{93.26} & 26.89 & 46.11 & 36.94 & 1.63 & 31.62\\
XLM-R & 21.92 & 42.55 & \textbf{75.16} & \textbf{87.60} & 69.79 & 86.13 & 92.26 & 34.52 & 51.12 & 44.81 & 0.81 & 54.88\\
AriaBERT & 16.49 & 37.98 & 69.70 & 82.71 & 63.61 & 80.71 & 89.08 & 22.68 & 41.37 & 32.89  & 0 & 40.93\\
\midrule
FaBERT & \textbf{33.33} & \textbf{55.87} & 75.04 & 87.34 & \textbf{70.33} & \textbf{86.50} & 90.02 & \textbf{35.85} & \textbf{53.51} & \textbf{45.36} & \textbf{2.45} & \textbf{61.39}\\
\midrule
Human & - & - & 80.3 &  88.3 & 74.9 & 85.6 & 96.80 & 85.5 & 86.97 & -  & - & -\\
\bottomrule
\end{tabular}
\end{adjustbox}
\caption{Performance Comparison in Question Answering} \label{tab:qa_results}
\end{table*}

\subsection{Question Answering}
To evaluate the question-answering capabilities of FaBERT, our experiments encompassed three datasets: ParsiNLU Reading Comprehension \cite{khashabi2021parsinlu}, PQuad \cite{darvishi2023pquad}, and PCoQA \cite{hemati2023pcoqa}. Each dataset is briefly introduced in the following sections. Table \ref{tab:qa_results} summarizes the performance of different models on each dataset.

\subsubsection*{ParsiNLU Reading Comprehension Dataset}
Reading Comprehension is one of the tasks introduced in the ParsiNLU benchmark and involves extracting a substring from a given context paragraph to answer a specific question. In order to create this dataset, they used Google’s Autocomplete API to mine questions deemed popular by users. Starting with a seed set of questions, they repeatedly queried previous questions to expand on the set and add more sophisticated ones. After filtering out invalid questions, native annotators then chose the pertinent text span from relevant paragraphs that provided the answer to each question.

% \begin{table}[htbp]
% \centering
% \begin{adjustbox}{width={0.3\textwidth},totalheight={\textheight},keepaspectratio}
%   \begin{tabular}{lcc}
%     \toprule
%     \textbf{Model} & \textbf{Exact Match} & \textbf{F1} \\
%     \midrule
%     ParsBERT & 22.10 & 44.89 \\
%     MBERT & 26.31 & 49.63 \\
%     XLM & 21.92 & 42.55 \\
%     AriaBERT & 16.49 & 37.98 \\
%     FaBERT & 33.33 & 55.87 \\
%     \bottomrule
%   \end{tabular}
% \end{adjustbox}
% \caption{Performance Comparison on Exact Match and F1}\label{tab:QA_Results}
% \end{table}
The evaluation of models on this dataset involves comparing the answers generated by the models to the provided ground truth answers. The main metrics used are the F1 score, which measures the overlap between the predicted and ground truth answers, and the exact match (EM) score, which checks if the predicted answers exactly match the ground truth answers. FaBERT scored +6.24\% higher in F1 compared to other models in the ParsiNLU Reading Comprehension task.

\subsubsection*{PQuAD: A Persian question answering dataset}
PQuAD is a large-scale, human-annotated question-answering dataset for the Persian language. It contains 80,000 questions based on passages extracted from Persian Wikipedia articles. The questions and their corresponding answers were generated through a crowdsourcing process, where crowdworkers were presented with passages and tasked with crafting questions and corresponding answers based on the provided content. Inspired by the structure of SQuAD 2.0 \cite{rajpurkar2018know}, PQuAD designates 25\% of its questions as unanswerable, adding extra complexity to the dataset and enhancing the evaluative challenge.

% \begin{table}[htbp]
% \centering
% \begin{adjustbox}{width={0.5\textwidth},totalheight={\textheight},keepaspectratio}
%   \begin{tabular}{lccccc}
%     \toprule
%     \textbf{Model} & \textbf{Exact Match} & \textbf{F1} & \textbf{HasAns EM} & \textbf{HasAns F1} & \textbf{NoAns EM/F1} \\
%     \midrule
%     ParsBERT & 74.41 & 86.89 & 68.97 & 85.34 & 91.79 \\
%     MBERT & 73.68 & 86.71 & 67.52 & 84.66 & 93.26 \\
%     XLM & 75.16 & 87.60 & 69.79 & 86.13 & 92.26 \\
%     AriaBERT & 69.70 & 82.71 & 63.61 & 80.71 & 89.08 \\
%     FaBERT & 74.78 & 87.24 & 69.90 & 86.29 & 90.28 \\
%     Human [pq] & 80.3 & 88.3 & 74.9 & 85.6 & 96.8 \\
%     \bottomrule
%   \end{tabular}
% \end{adjustbox}
% \caption{QA Performance Comparison on Exact Match, F1, HasAns EM, HasAns F1, and NoAns EM/F1}\label{tab:QA_Results}
% \end{table}

In this dataset, in addition to F1 and EM scores, the evaluation can be broken down into subsets of questions that have answers (HasAns) and those that do not have answers (NoAns). By considering these metrics, the performance of different models can be compared and analyzed to determine their effectiveness in answering questions or abstaining from answering. The authors also provided an estimation of human performance by asking a group of crowdworkers to answer a subset of questions. Both FaBERT and XLM-R demonstrate remarkable capabilities in question answering, achieving a comparable F1 score performance. However, XLM-R slightly outperforms FaBERT in this aspect.

\subsubsection*{PCoQA: Persian Conversational Question Answering Dataset}
PCoQA is the first dataset designed for answering conversational questions in Persian. It comprises 870 dialogs and over 9,000 question-answer pairs sourced from Wikipedia articles.
% PCoQA is the first-ever dataset for answering conversational questions in Persian. Comprising 870 dialogs and over 9,000 question-answer pairs sourced from Wikipedia articles, this dataset is structured around an information-seeking dialogue. 
In this task, contextually connected questions are posed about a given document, and models are required to respond by extracting relevant information from given paragraphs.
% Contextually connected questions are posed about a given document, and the answerer responds by extracting spans from the text.
This dataset provides a suitable context for assessing the model's performance in Persian conversational question answering, similar to the English dataset CoQA \cite{reddy2019coqa}.

For the PCoQA dataset, in addition to F1 and EM scores, two variants of human equivalence score (HEQ) are suggested by the authors. HEQ-Q measures the percentage of questions for which system F1 exceeds or matches human F1, and HEQ-M quantifies the number of dialogs for which the model achieves a better overall performance compared to the human. FaBERT outperformed other models with +2.55\% higher F1 score, handling both answerable and unanswerable questions well. Additionally, the PCoQA dataset proves to be challenging, with all models scoring noticeably lower than humans.
% \begin{table}[htbp]
% \centering
% \begin{adjustbox}{width={0.5\textwidth},totalheight={\textheight},keepaspectratio}
%   \begin{tabular}{lcccccc}
%     \toprule
%     \textbf{Model} & \textbf{Exact Match} & \textbf{F1} & \textbf{HEQ-Q} & \textbf{HEQ-M} & \textbf{HEQ-D} & \textbf{NoAns EM/F1} \\
%     \midrule
%     ParsBERT & 31.17 & 50.96 & 41.07 & 0.81 & 0 & 48.83 \\
%     M-BERT & 46.11 & 36.94 & 1.63 & 0.81 & 31.62 & - \\
%     XLM-R & 34.52 & 51.12 & 44.81 & 0.81 & 0.81 & 54.88 \\
%     AriaBERT & 22.68 & 41.37 & 32.89 & 0 & 0 & 40.93 \\
%     Human [PcoQA] & 85.50 & 86.97 & - & - & - & - \\
%     FaBERT & 35.85 & 53.51 & 45.36 & 2.45 & 0.0 & 61.39 \\
%     \bottomrule
%   \end{tabular}
% \end{adjustbox}
% \caption{Performance Comparison on Exact Match, F1, HEQ-Q, HEQ-M, HEQ-D, and NoAns EM/F1}\label{tab:QA_HEQ_Results}
% \end{table}

\subsection{Vocabulary Impact on Input Length}
To evaluate the impact of FaBERT's chosen vocabulary size on its effective maximum input length, a comparative analysis was conducted across datasets with longer sentences, including MirasOpinion, FarsTail, ParsiNLU Reading Comprehension, and PQuAD. The objective was to examine how different tokenizers, including the one trained for FaBERT, influence the number of tokens in each input sentence.

Table \ref{tab:Median_Token} provides a summary of median token counts across the aforementioned datasets. Both multilingual models faced challenges due to the lack of sufficient Persian tokens in their vocabularies, potentially impacting their performance on longer inputs due to loss of information. ParsBERT's tokenizer yields the most compact sequences, closely followed by FaBERT. An interesting observation arises in the PQuAD dataset, where ParsBERT outperforms, likely attributed to PQuAD's reliance on Wikipedia, a significant component of ParsBERT's pre-training data.

Overall, FaBERT's tokenizer, despite having a vocabulary size half that of ParsBERT, demonstrated a comparable level of compression. The detailed boxplots for each dataset are available in Appendix \ref{sec:appendix_tokenizer}.

\begin{table}[htbp]
\centering
\begin{adjustbox}{width={0.5\textwidth},totalheight={\textheight},keepaspectratio}
\begin{tabular}{lcccc}
    \toprule
    \textbf{Tokenizer} & \textbf{MirasOpinion} & \textbf{FarsTail} & \textbf{ParsiNLU RC} & \textbf{PQuAD} \\
    \midrule
    ParsBERT & 27 & 58 & 113.5 & 160 \\
    mBERT & 44 & 85 & 165 & 235 \\
    XLM-R & 34 & 74 & 142.5 & 210 \\
    AriaBERT & 28 & 66 & 130 & 207 \\
    FaBERT & 28 & 62 & 119.5 & 189 \\
    \bottomrule
\end{tabular}
\end{adjustbox}
\caption{Median Token Count Yielded by Different Tokenizers}\label{tab:Median_Token}
\end{table}

\section{Conclusion}\label{sec:conclusion}
In this paper, we pre-trained FaBERT, a BERT-base model from scratch exclusively on the cleaned HmBlogs corpus, consisting solely of raw texts from Persian blogs. Notably, our model's smaller vocabulary size resulted in a more compact overall size compared to competitors. FaBERT performed exceptionally well in 12 different datasets, outperforming competitors in nine of them. In the remaining tasks where it did not secure the top position, it consistently ranked among the top performers, closely following the highest-performing model. Our results indicate that clean texts with diverse writing styles, both formal and informal, found in Persian blogs can significantly contribute to the high-quality pre-training of language models, including BERT. The effectiveness of the Hmblogs corpus in the performance of our BERT model in downstream tasks demonstrates its potential for being used in pre-training both language models and large language models alongside other relevant Persian corpora.

%------------------------------------------------------------------------------------------------------------------------------------------------------------------------------------------------------------------
% \newpage
\bibliography{references}
\bibliographystyle{acl_natbib}

%------------------------------------------------------------------------------------------------------------------------------------------------------------------------------------------------------------------
% \newpage

\newpage
\section*{Appendix For "FaBERT: Pre-training BERT on Persian Blogs"}
\label{sec:appendix_title}

\appendix

    \section{Fine-tuning Hyperparameters}
    \label{sec:appendix_hyper}
    
    The hyperparameters employed for fine-tuning the models on each dataset, along with the respective train/validation/test split sizes, are outlined in Table \ref{tab:dataset_info}. For the ParsiNLU benchmark, we adhered to the predefined hyperparameters in the ParsiNLU source code.

% \subsection{Appendix A}\label{sec:appendixA}
% The hyperparameters employed for fine-tuning the models on each dataset, along with the respective train/validation/test split sizes, are outlined in Table \ref{tab:AppendixA}.

\begin{table*}[htbp]
\centering
\begin{adjustbox}{width={\textwidth},totalheight={\textheight},keepaspectratio}
\begin{tabular}{lccccccccc}
\toprule
\textbf{Datasets} & \textbf{Train} & \textbf{Validation} & \textbf{Test} & \textbf{Number of Labels} & \textbf{Metrics} & \textbf{Learning Rate} & \textbf{Batch Size} & \textbf{Epochs} & \textbf{Warmup}\\
\midrule
DeepSentiPers & 6320 & 703 & 1854 & 3 & Macro F1 & 2e-5, 3e-5, 5e-5 & 8,16 & 3, 7 & 0, 0.2 \\
MirasOpinion & 75094 & 9387 & 9387 & 3 & Macro F1 & 2e-5, 3e-5, 5e-5 & 8,16 & 1 & 0, 0.2\\
MirasIrony & 2352 & 295 & 294 & 2 & Accuracy & 2e-5, 3e-5, 5e-5 & 8,16 & 3, 5 & 0, 0.2\\
\bottomrule
PQuAD & 63994 & 7976 & 8002 & - & Micro F1 & 2e-5, 3e-5, 5e-5 & 8,16 & 2 & 0, 0.2\\
PCoQA & 6319 & 1354 & 1354 & - & Micro F1 & 3e-5, 5e-5 & 8,16 & 3, 7 & 0, 0.2\\
ParsiNLU RC & 600 & 125 & 575 & - & Micro F1 & 3e-5, 5e-5 & 4 & 3, 7 & 0\\
\bottomrule
SBU-NLI & 3248 & 361 & 401 & 3 & Micro F1 & 2e-5, 3e-5, 5e-5 & 8,16 & 3, 7 & 0, 0.2\\
FarsTail & 7266 & 1564 & 1537 & 3 & Micro F1 & 2e-5, 3e-5, 5e-5 & 8,16 & 3, 7 & 0, 0.2\\
ParsiNLU QP & 1830 & 898 & 1916 & 2 & Micro F1 & 3e-5, 5e-5 & 8,16 & 3, 7 & 0\\
\bottomrule
PEYMA & 8029 & 926 & 1027 & - & Macro F1 & 2e-5, 3e-5, 5e-5 & 8,16 & 3, 7 & 0, 0.2 \\
MultiCoNER v2 & 16321 & 855 & 219168 & - & Micro F1 & 2e-5, 3e-5, 5e-5 & 8,16 & 3, 7 & 0, 0.2\\
ParsTwiNER & 6418 & 447 & 304 & - & Micro F1 & 2e-5, 3e-5, 5e-5 & 8,16 & 3, 7 & 0, 0.2\\
\bottomrule
\end{tabular}
\end{adjustbox}
\caption{Dataset Split Sizes and Fine-Tuning Hyperparameters}
\label{tab:dataset_info}
\end{table*}
%------------------------------------------------------------------------------------------------------------------------------------------------------------------------------------------------------------------

    \section{Detailed NER Results}
    \label{sec:appendix_ner}
    
    Tables~\ref{tab:entity_comparison_PEYMA},~\ref{tab:entity_comparison_ParsTwiNER}, and~\ref{tab:entity_comparison_MultiCoNER} present F1 scores for entities in PEYMA, MultiCoNER v2, and ParsTwiNER datasets, providing a model comparison for each entity. For instance, In MultiCoNER v2, FaBERT excels in recognizing medical entities, and ParsBERT is better at identifying creative works.

\begin{table*}[htbp]
\centering
\begin{adjustbox}{width={0.7\textwidth},totalheight={\textheight},keepaspectratio}
\begin{tabular}{lccccc|c}
\toprule
\textbf{Entity Type} & \textbf{FaBERT} & \textbf{ParsBERT} & \textbf{AriaBERT} & \textbf{mBERT} & \textbf{XLM-R} & \textbf{Support}\\
\midrule
Date & 89.16 & 85.65 & 85.11 & 84.56 & 86.73 & 208\\
Location & 91.95 & 91.73 & 91.46 & 90.25 & 92.42 & 595\\
Currency & 94.34 & 94.34 & 83.64 & 90.57 & 96.15 & 26\\
Organization & 88.24 & 89.37 & 86.38 & 84.83 & 87.25 & 667\\
Percent & 98.63 & 98.63 & 93.33 & 97.14 & 94.74 & 36\\
Person & 95.45 & 95.29 & 94.6 & 90.1 & 95.75 & 434\\
Time & 96.97 & 91.43 & 96.97 & 76.47 & 94.12 & 16\\
\midrule
\textbf{Micro Average} & 91.39 & 91.24 & 89.76 & 87.84 & 90.91 & 1982\\
\textbf{Macro Average} & 93.53 & 92.35 & 90.21 & 87.7 & 92.45 & 1982\\
\textbf{Weighted Average} & 91.37 & 91.23 & 89.75 & 87.81 & 90.92 & 1982\\
\bottomrule
\end{tabular}
\end{adjustbox}
\caption{Comparison of F1 Scores for Each Entity Type in PEYMA}\label{tab:entity_comparison_PEYMA}
\end{table*}

\begin{table*}[htbp]
\centering
\begin{adjustbox}{width={0.7\textwidth},totalheight={\textheight},keepaspectratio}
  \begin{tabular}{lccccc|c}
    \toprule
    \textbf{Entity Type} & \textbf{FaBERT} & \textbf{ParsBERT} & \textbf{AriaBERT} & \textbf{mBERT} & \textbf{XLM-R} & \textbf{Support} \\
    \midrule
    Event & 0.5714 & 0.4444 & 0.4118 & 0.4865 & 0.2308 & 14 \\
    Location & 0.8281 & 0.8414 & 0.7991 & 0.7802 & 0.8088 & 221 \\
    Nation & 0.9 & 0.7385 & 0.7246 & 0.7123 & 0.7397 & 30 \\
    Organization & 0.7364 & 0.6966 & 0.6691 & 0.6462 & 0.7126 & 129 \\
    Person & 0.9344 & 0.8893 & 0.8745 & 0.8216 & 0.8629 & 244 \\
    Political Group & 0.6364 & 0.6667 & 0.7442 & 0.7 & 0.8 & 22 \\
    \midrule
    \textbf{Micro Average} & 0.8222 & 0.8113 & 0.7853 & 0.756 & 0.795 & 660 \\
    \textbf{Macro Average} & 0.7301 & 0.7128 & 0.7039 & 0.6911 & 0.6925 & 660 \\
    \textbf{Weighted Average} & 0.8238 & 0.8119 & 0.7881 & 0.7573 & 0.7943 & 660 \\
    \bottomrule
  \end{tabular}
\end{adjustbox}
\caption{Comparison of F1 Scores for Each Entity Type in ParsTwiNER}\label{tab:entity_comparison_ParsTwiNER}
\end{table*}

\setlength\dashlinedash{4pt}
\setlength\dashlinegap{8pt}

\begin{table*}[htbp]
\centering
\begin{adjustbox}{width={0.7\textwidth},totalheight={\textheight},keepaspectratio}
  \begin{tabular}{lccccc|c}
    \toprule
    \textbf{Entity Type} & \textbf{FaBERT} & \textbf{ParsBERT} & \textbf{AriaBERT} & \textbf{mBERT} & \textbf{XLM-R} & \textbf{Support} \\
    \midrule
    AerospaceManufacturer & 0.7325 & 0.7127 & 0.7196 & 0.6269 & 0.638 & 1030 \\
    ORG & 0.5809 & 0.5832 & 0.5348 & 0.5479 & 0.5325 & 18532 \\
    MusicalGRP & 0.6282 & 0.6597 & 0.59 & 0.613 & 0.5954 & 4668 \\
    PrivateCorp & 0.3822 & 0.4033 & 0.3851 & 0.2605 & 0.1749 & 148 \\
    CarManufacturer & 0.6511 & 0.7031 & 0.6631 & 0.6291 & 0.6147 & 2085 \\
    PublicCorp & 0.6109 & 0.6377 & 0.5819 & 0.5439 & 0.562 & 5926 \\
    SportsGRP & 0.8159 & 0.8174 & 0.8012 & 0.8046 & 0.7949 & 6418 \\
    \hdashline
    Medication/Vaccine & 0.7067 & 0.6837 & 0.6342 & 0.6324 & 0.6582 & 4405 \\
    MedicalProcedure & 0.6307 & 0.5965 & 0.5592 & 0.4904 & 0.5471 & 2132 \\
    AnatomicalStructure & 0.6079 & 0.5827 & 0.5151 & 0.4824 & 0.4978 & 3940 \\
    Symptom & 0.5656 & 0.5368 & 0.4671 & 0.4217 & 0.4109 & 821 \\
    Disease & 0.646 & 0.6256 & 0.5737 & 0.5264 & 0.5652 & 3989 \\
    \hdashline
    Artist & 0.7384 & 0.7347 & 0.6936 & 0.7122 & 0.7155 & 51617 \\
    Politician & 0.5786 & 0.6056 & 0.534 & 0.5213 & 0.5141 & 19760 \\
    Scientist & 0.3328 & 0.3669 & 0.2952 & 0.2615 & 0.2625 & 3278 \\
    SportsManager & 0.606 & 0.6232 & 0.5376 & 0.4332 & 0.4494 & 3009 \\
    Athlete & 0.5796 & 0.5992 & 0.5356 & 0.5119 & 0.5357 & 12551 \\
    Cleric & 0.5707 & 0.5535 & 0.4875 & 0.4627 & 0.4332 & 4526 \\
    OtherPER & 0.4254 & 0.4225 & 0.3544 & 0.3647 & 0.3449 & 21127 \\
    \hdashline
    Clothing & 0.3912 & 0.3375 & 0.3293 & 0.2054 & 0.2716 & 239 \\
    Drink & 0.5244 & 0.5683 & 0.5483 & 0.4646 & 0.5041 & 631 \\
    Food & 0.6063 & 0.5971 & 0.574 & 0.4788 & 0.5591 & 3580 \\
    Vehicle & 0.5388 & 0.5388 & 0.5171 & 0.4659 & 0.4952 & 2865 \\
    OtherPROD & 0.5851 & 0.5843 & 0.5453 & 0.5109 & 0.5233 & 10897 \\
    \hdashline
    ArtWork & 0.0919 & 0.1085 & 0.1057 & 0.1077 & 0.0691 & 100 \\
    WrittenWork & 0.5561 & 0.5541 & 0.5028 & 0.5006 & 0.5079 & 13530 \\
    VisualWork & 0.7447 & 0.7463 & 0.7095 & 0.7445 & 0.7523 & 25054 \\
    Software & 0.6448 & 0.6586 & 0.5991 & 0.5913 & 0.5911 & 8058 \\
    MusicalWork & 0.5408 & 0.5714 & 0.5239 & 0.5492 & 0.545 & 6292 \\
    \hdashline
    Facility & 0.5673 & 0.5671 & 0.5283 & 0.5317 & 0.5347 & 11393 \\
    Station & 0.7997 & 0.7863 & 0.7812 & 0.784 & 0.781 & 2532 \\
    HumanSettlement & 0.7608 & 0.7676 & 0.7517 & 0.7658 & 0.7647 & 55741 \\
    OtherLOC & 0.37 & 0.3348 & 0.3413 & 0.2965 & 0.2376 & 1241 \\
    \midrule
    \textbf{Micro Average} & 0.6451 & 0.6517 & 0.6081 & 0.6108 & 0.6145 & 312115 \\
    \textbf{Macro Average} & 0.5792 & 0.5809 & 0.54 & 0.5104 & 0.5147 & 312115 \\
    \textbf{Weighted Average} & 0.6491 & 0.6531 & 0.6101 & 0.6111 & 0.6131 & 312115 \\
    \bottomrule
  \end{tabular}
\end{adjustbox}
\caption{Comparison of F1 Scores for Each Entity Type in MultiCoNER v2} \label{tab:entity_comparison_MultiCoNER}
\end{table*}

%------------------------------------------------------------------------------------------------------------------------------------------------------------------------------------------------------------------

    \section{Tokenizer Comparison Figures}
    \label{sec:appendix_tokenizer}
    
\begin{figure}[!h]
\centering
\includegraphics[width=0.95\linewidth]{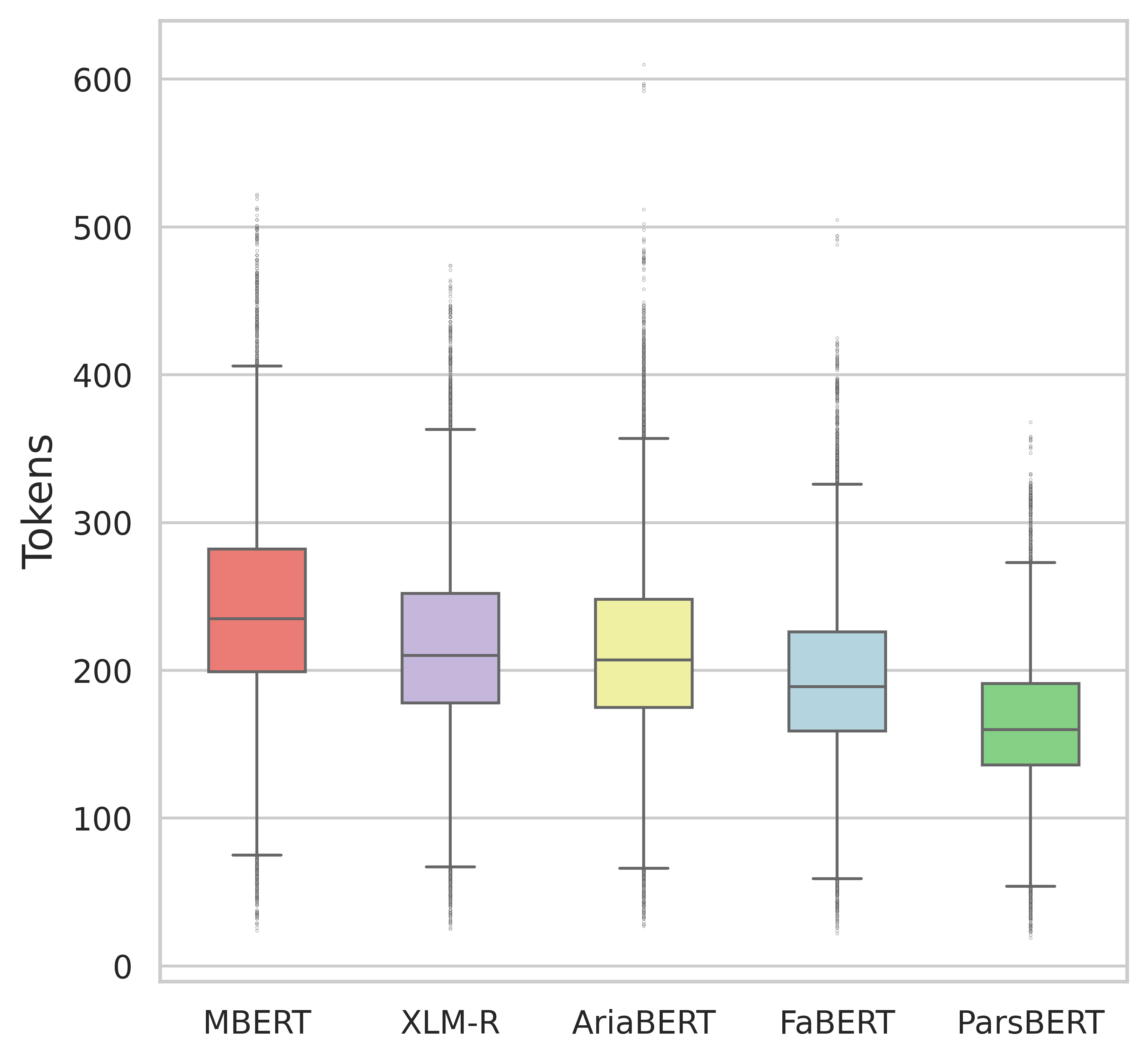}
\caption{Token count distribution across tokenizers for the PQuAD dataset}
\label{fig:pquad}
\end{figure}

\begin{figure}
\centering
\includegraphics[width=0.95\linewidth]{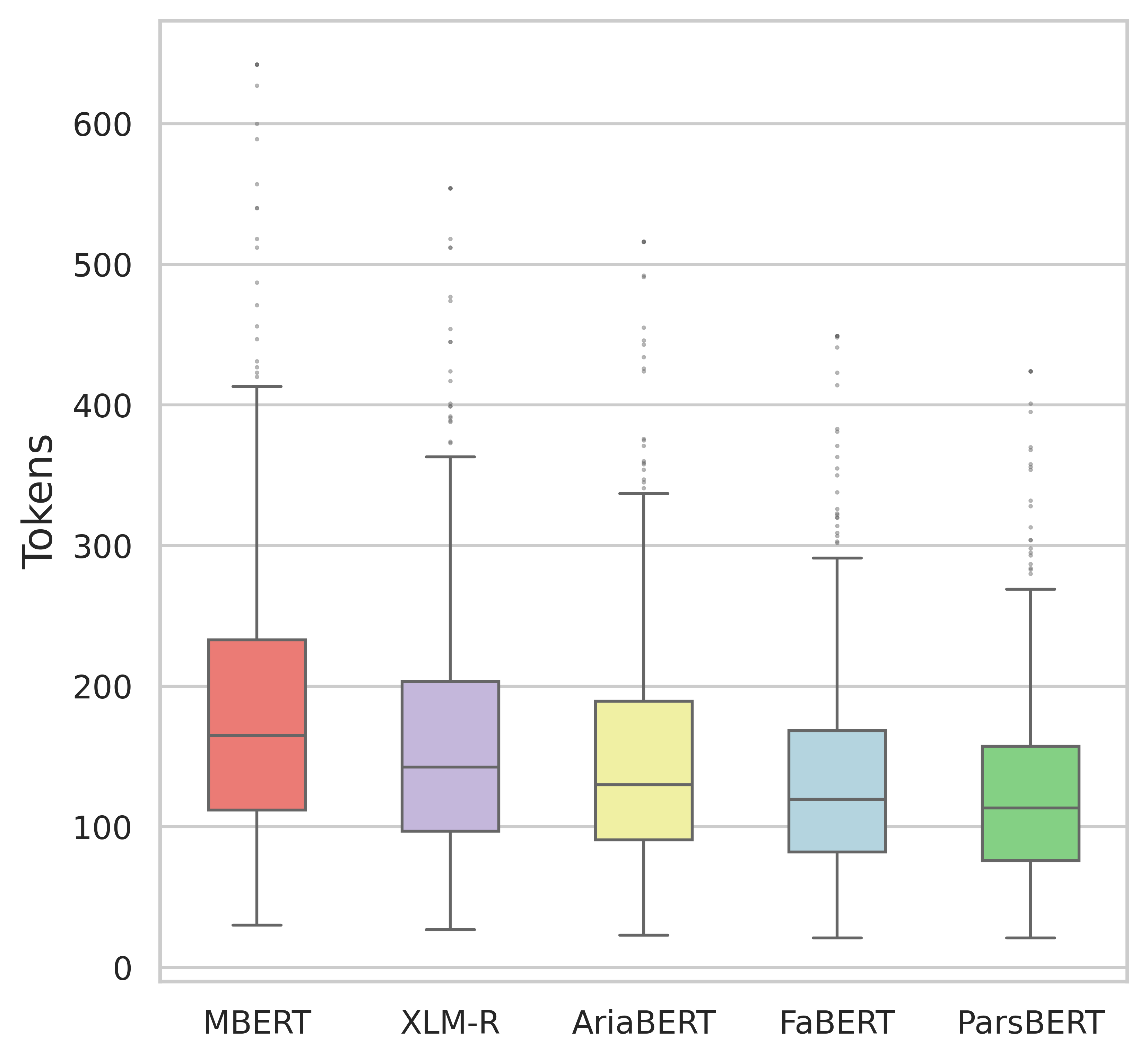}
\caption{Token count distribution across model tokenizers for the ParsiNLU Reading Comprehension dataset}
\label{fig:parsinlu}
\end{figure}

\begin{figure}
\centering
\includegraphics[width=0.95\linewidth]{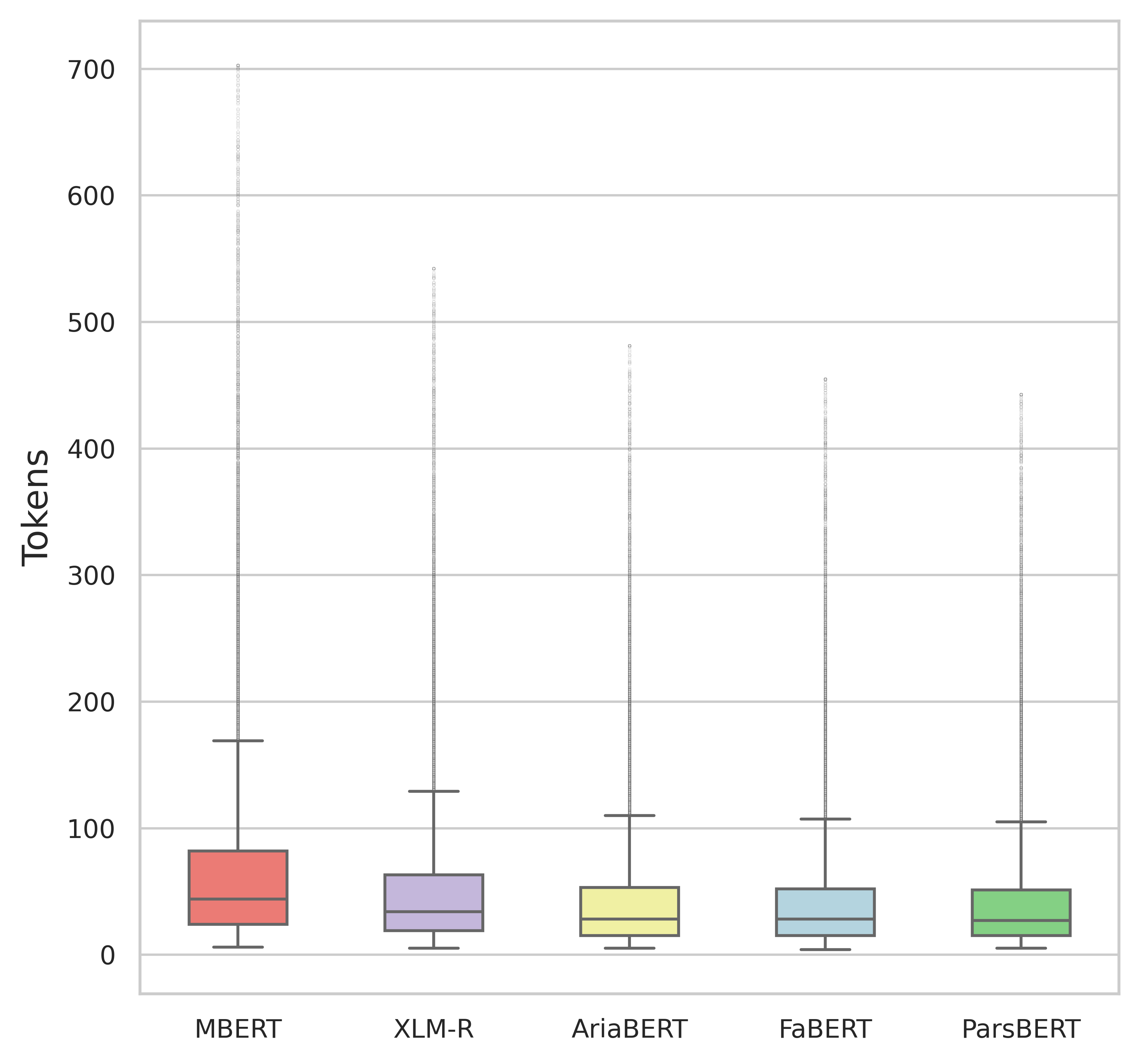}
\caption{Token count distribution across tokenizers for the MirasOpinion dataset}
\label{fig:mirasopinion}
\end{figure}

\begin{figure}
\centering
\includegraphics[width=0.95\linewidth]{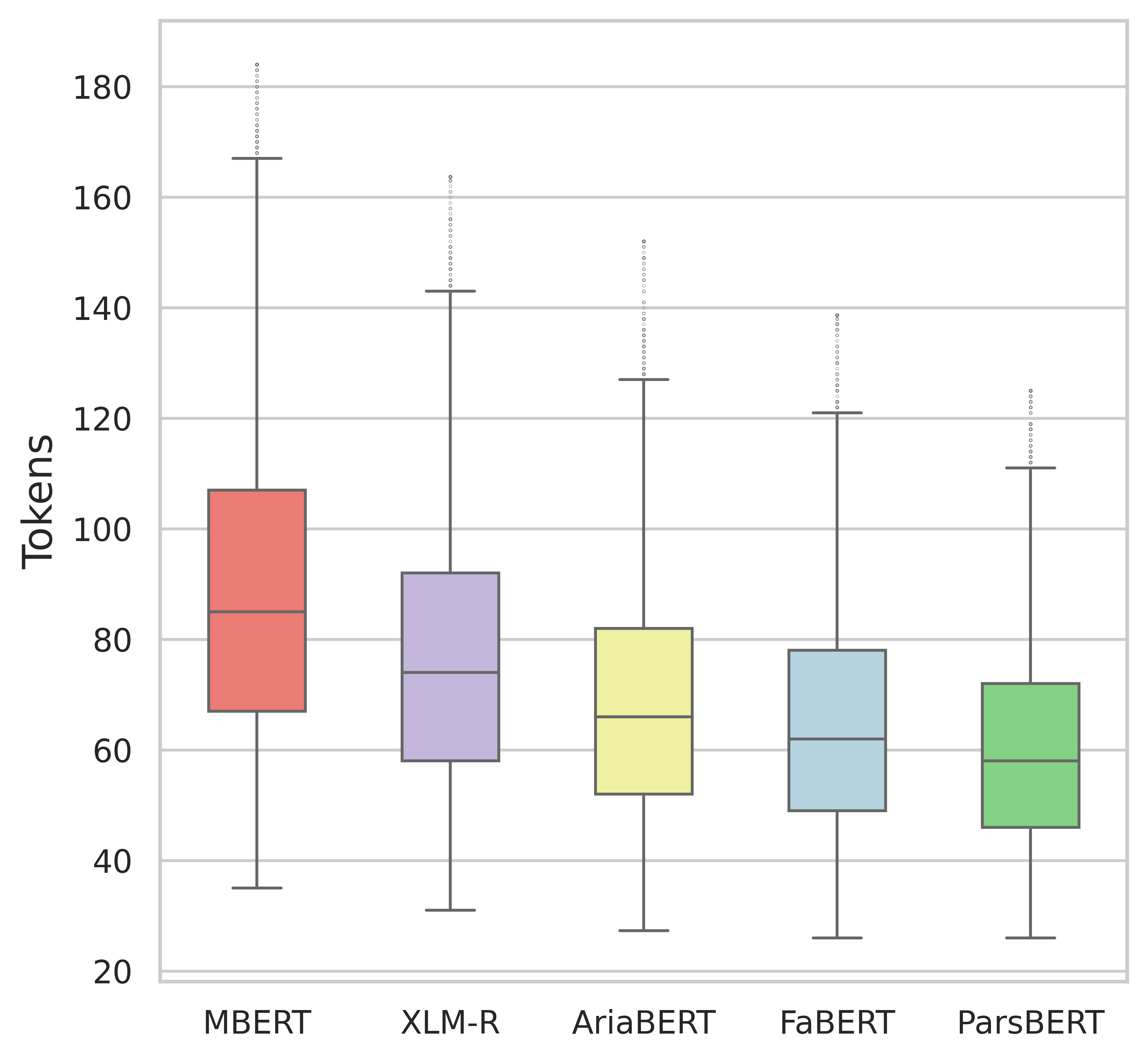}
\caption{Token count distribution across tokenizers for the FarsTail dataset}
\label{fig:farstail}
\end{figure}

    Figures~\ref{fig:pquad},~\ref{fig:parsinlu},~\ref{fig:mirasopinion}, and~\ref{fig:farstail} illustrate the distribution of token counts for each model's tokenizer across the following datasets: PQuAD, ParsiNLU Reading Comprehension, MirasOpinion, and FarsTail. These boxplots provide a visual representation of the variation in token counts for each model.

\end{document}